\documentclass[journal,twoside,web]{ieeecolor}
\usepackage{mathtools}
\usepackage{multirow}
\usepackage{algorithm}  
\usepackage{algpseudocode}  
\usepackage{comment}
\usepackage{stfloats}
\usepackage{siunitx}
\usepackage{booktabs}
\usepackage{tablefootnote}
\usepackage{siunitx}
\usepackage{jsen}
\usepackage{cite}
\usepackage{amsmath,amssymb,amsfonts}
\usepackage{graphicx}
\usepackage{textcomp}
\usepackage{wrapfig}
\def\BibTeX{{\rm B\kern-.05em{\sc i\kern-.025em b}\kern-.08em
    T\kern-.1667em\lower.7ex\hbox{E}\kern-.125emX}}
\definecolor{abstractbg}{rgb}{0.89804,0.94510,0.83137}
\newcommand{\eg}{\textit{e.g.,}}
\newcommand{\ie}{\textit{i.e.,}}

\begin{document}
\title{Feature Pyramid biLSTM: Using Smartphone Sensors for Transportation Mode Detection}
\author{Qinrui Tang, Hao Cheng, \IEEEmembership{Member, IEEE}
\thanks{This research is supported by project VMo4Orte. }
\thanks{Qinrui Tang is with Institute of Transportation Systems, German Aerospace Center (DLR), Berlin, Germany (e-mail: qinrui.tang@dlr.de; qinruitang@gmail.com).}
\thanks{Hao Cheng is with ITC Faculty Geo-Information Science and Earth Observation, University of Twente, Enschede, the Netherlands (e-mail: h.cheng-2@utwente.nl).}
}

\IEEEtitleabstractindextext{%
\fcolorbox{abstractbg}{abstractbg}{%
\begin{minipage}{\textwidth}%
\begin{wrapfigure}[13]{r}{3in}%
\includegraphics[width=3in]{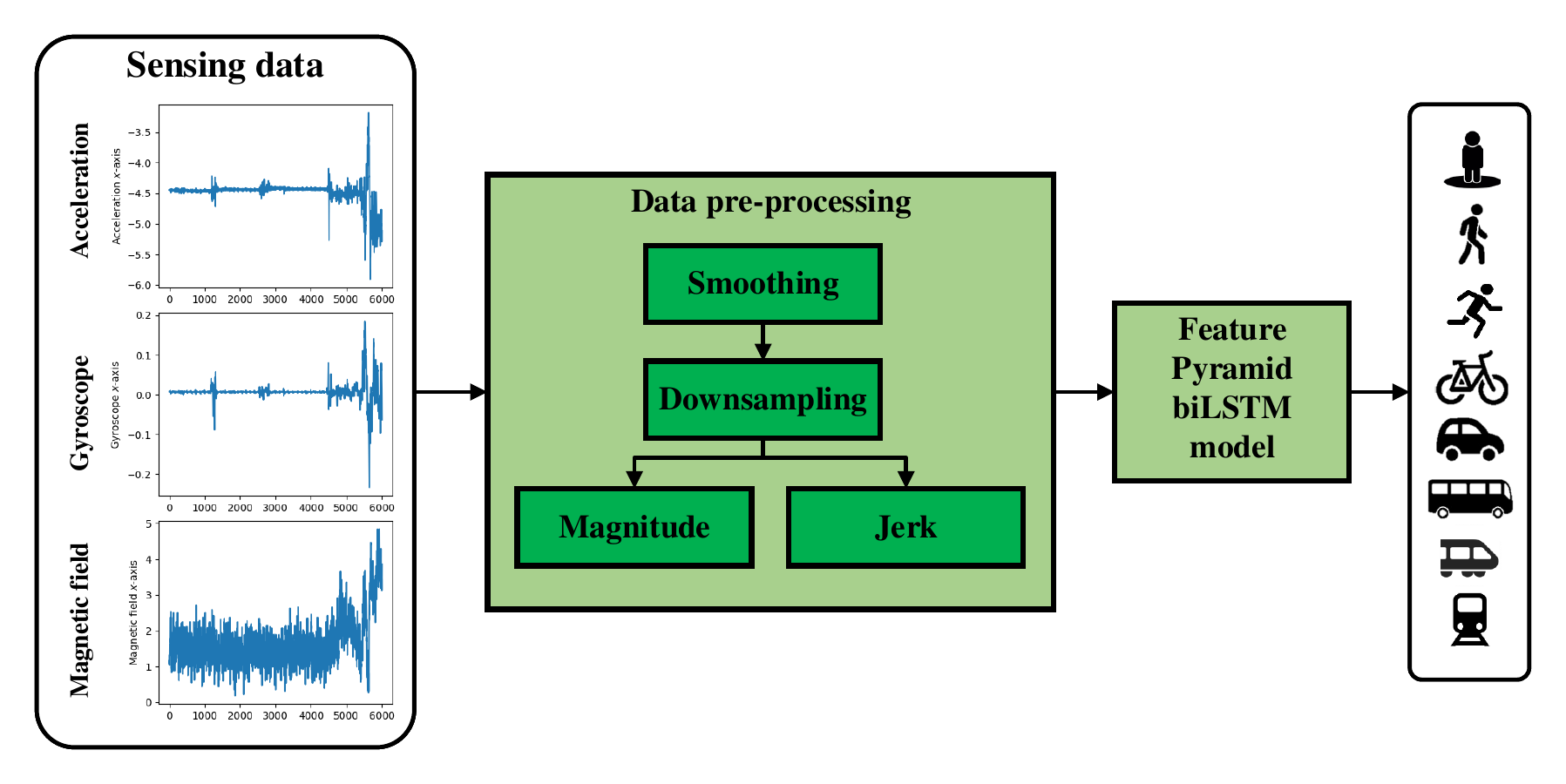}%
\end{wrapfigure}%
\begin{abstract}
The widespread utilization of smartphones has provided extensive availability to Inertial Measurement Units, providing a wide range of sensory data that can be advantageous for the detection of transportation modes. 
The objective of this study is to propose a novel end-to-end approach to effectively explore a reduced amount of sensory data collected from a smartphone to achieve accurate mode detection in common daily traveling activities.
Our approach, called Feature Pyramid biLSTM (FPbiLSTM), is characterized by its ability to reduce the number of sensors required and processing demands, resulting in a more efficient modeling process without sacrificing the quality of the outcomes than the other current models.
FPbiLSTM extends an existing CNN biLSTM model with the Feature Pyramid Network, leveraging the advantages of both shallow layer richness and deeper layer feature resilience for capturing temporal moving patterns in various transportation modes. 
It exhibits an excellent performance by employing the data collected from only three out of seven sensors, \ie~accelerometers, gyroscopes, and magnetometers, in the 2018 Sussex-Huawei Locomotion (SHL) challenge dataset, attaining a noteworthy accuracy of 95.1\% and an $F_{1}$-score of 94.7\% in detecting eight different transportation modes. 

\end{abstract}

\end{minipage}}}

\maketitle

\section{Introduction}
\label{sec:intro}
In the last decade, smartphones have gained widespread prevalence in daily existence, hence facilitating an unparalleled degree of accessibility to Inertial Measurement Units (IMUs) that are included within these devices. The utilization of IMUs comprising of \eg~Microelectromechanical Systems (MEMS) accelerometers, gyroscopes, and magnetometers, has the potential to provide a wide range of sensory data that can be advantageous for various applications. These applications include motion tracking, indoor positioning, and transportation mode detection.

This study focuses on the issue of detecting different transportation modes using smartphone sensory data. This task can be viewed as a classification problem encompassing several categories such as walking, biking, driving a car, and others.
Obtaining accurate information about the user's chosen mode of transportation is crucial for enhancing decision-making procedures and formulating effective strategies for urban transportation planning \cite{Xiao2012,liang2019deep}. 
In addition, it is crucial to comprehend the transportation mode preferences of a passenger in order to provide tailored advertising strategies and optimize the efficiency of transportation surveys. 
The traditional approach to conducting interviews, due to its high costs and time requirements, can potentially be replaced with a more efficient and automated system for collecting and categorizing data. 
As smartphones with various IMUs are commonly carried by the users while traveling, they are a perfect type of devices to be used for transportation mode detection.
Moreover, the detection of transportation mode can assist in approximating a user's overall location once the appropriate mode of transportation has been determined.


The problem of mode detection has been tackled through the utilization of both traditional machine learning methods and deep learning approaches. In the context of traditional machine learning, there is a frequent requirement for feature extraction and domain expertise. The inclusion of these conditions may result in an increase in workloads and could potentially restrict the adaptation of the technique to similar issues with modest differences in sensor configurations, hence presenting a drawback in their applications. In contrast, deep learning has facilitated a multitude of researchers in attaining elevated levels of accuracy or $F_1$-scores throughout the evaluation of their models. Nevertheless, these models commonly employ a wide range of sensors including accelerometers, gyroscopes, magnetometers, linear accelerometers, gravity sensors, orientation sensors, and ambient pressure sensors. The presence of a larger quantity of sensors can accidentally result in an increased demand for greater capacity for data storage, additional resources for data preprocessing, and perhaps heightened requirements for model training resources.

To comply with computational limits, utilizing a reduced number of input signals for transportation mode detection is more desirable for an economical deployment.
The early works, \eg~\cite{ito2018application} applied a CNN-based model that takes only the spectrogram from both accelerometer and magnetometer measurements, and \cite{widhalm2018top} employed a multilayer perception-based model with an additional magnetometer measurement. 
However, their performances are noticeably inferior to the other models that leverage the measurements from more sensors. 
Moreover, Tang et al. \cite{tang2022} attempted to propose a more advanced network framework to compensate the limited amount of measurements.
The approach involves the integration of Convolutional Neural Network (CNN) and bidirectional Long Short-Term Memory (biLSTM) models, and utilizes accelerometer and magnetometer measurements for the detection. 
It demonstrates on-par levels of accuracy and $F_1$-scores as the other methods that utilize a greater number of input signals.
In this paper, we follow the constrain of using a reduced number of sensory measurements and seek for a more effective end-to-end approach for transportation mode detection.
Namely, this study introduces a novel approach called Feature Pyramid biLSTM (FPbiLSTM), which is inspired by the Feature Pyramid Network (FPN) proposed by Lin et al. \cite{lin2017featurepyramid}. 
The FPN leverages the abundant information present in lower levels and the robust features extracted from higher layers, resulting in enhanced model performance. 
Transportation modes can differ significantly in travel speeds, such as walking and taking a car, while they can also present very similar motion patterns, such as riding in a subway or a train.
Hence, this pyramid architecture is exploited to learn the temporal patterns from the measurements at different time granularity, in order to distinguish different transportation modes from both heterogeneous and homogeneous motion patterns.
This assertion is substantiated in the experimental phase of our study by improving the $F_1$-score from 90.4\% to 94.2\% when compared to that of \cite{tang2022} under the same setting.
Moreover, our model is further enhanced by incorporating a feature selection technique that includes the additional readings from gyroscopes. 
Despite the increased complexity of the network structure and the addition of an extra sensor compared to \cite{tang2022}, FPbiLSTM remains a very efficient and lightweight end-to-end model. 

The \textit{main contributions} of this study are as follows.
\begin{itemize}
\item We propose a novel Feature Pyramid biLSTM model that encompasses the utilization of a reduced number of sensors and computational resources in comparison to earlier research efforts for transportation mode detection.
\item Our model is end-to-end as it takes as input the raw data and requires no extra feature extraction, post-processing, or ensembling techniques to boost the detection performance.
\item It exhibits an excellent performance by employing the data collected from only accelerometers, gyroscopes, and magnetometers, in the 2018 Sussex-Huawei Locomotion (SHL) challenge dataset, achieving an accuracy of 95.1\% and an $F_{1}$-score of 94.7\% in detecting eight different transportation modes. 
\end{itemize}


\section{Related work}

\subsection{Sensors used for transportation mode detection}
Given the ubiquity of smartphones, many researchers are exploring the potential of using smartphone sensory data to determine traffic modes. 
This data can be broadly split into two categories: motion-based and location-based \cite{yu2014big} sensory data. 
Motion-based data arises from devices like accelerometers, gyroscopes, magnetometers, linear accelerometers, gravity sensors, orientation (quaternions), and ambient pressure. 
While, location-based data typically stems from GPS or GNSS. 

In the motion-based applications, it is commonly observed that the utilization of a greater number of sensors tends to yield improved outcomes in terms of classification performance. 
However, the types of sensors used in cutting-edge approaches vary.
For example, the authors of \cite{gjoreski2020classical,janko2018new,janko2019cross,choi2019embracenet,zhu2020densenetx,kalabakov2020tackling} conducted studies utilizing data collected from seven distinct sensors, specifically the accelerometer, gyroscope, magnetometer, linear accelerometer, gravity, orientation (quaternions), and ambient pressure. These studies yielded highly favorable classification outcomes, with \cite{gjoreski2020classical} achieving an impressive $F_{1}$-score of 94.9\%. 
To reduce the storage and computational costs, there are works that prioritize the advancement of streamlined models that incorporate a reduced number of sensors with some sacrifice of classification performance. 
The study conducted by \cite{qin2019toward} employed a total of four sensors, namely the linear accelerometer, gyroscope, magnetometer, and pressure sensor. 
\cite{fang2017learning} futher reduced the number of sensors to three, specifically accelerometer, magnetometer, and gyroscope readings.  
Furthermore, \cite{ito2018application} employed a combination of only accelerometer and gyroscope sensors in their study, and \cite{tang2022} similarly utilized two sensors, namely accelerometer and magnetometer. 
In the single sensor case, Liang et al \cite{liang2019deep} exclusively utilized data obtained from the accelerometer. 
In general, the three most commonly utilized sensors in terms of their application are accelerometer, magnetometer, and gyroscope data.
In this paper, we also focus our approach on these three commonly used sensors and attempt to achieve the same level of performance as that of using more sensors.

Models that utilize location-based data have also demonstrated successful results. The sensors utilized in the acquisition of location-based data include GPS position, GPS reception, WiFi, and cellular technology. The studies conducted by \cite{saha2021empirical,balabka2021human} have yielded favorable outcomes when employing the aforementioned four sensors. GNSS is another option to be used \cite{munoz2023advanced}. Nevertheless, it is important to note that the overall efficacy of these sensors is typically lower compared to data obtained using motion-based methodologies.
In this work, our approach is focused on the utilization of motion-based data for transportation mode detection.

\subsection{Traditional machine learning methods}

Traditional machine learning methods have exhibited significant efficacy in tackling the problem of transportation mode detection, frequently resulting in a high level of accuracy. 
For example, efficient algorithms, \eg~XGBoost \cite{janko2018new,kalabakov2020tackling,zhu2020densenetx} and Random Forest \cite{antar2018comparative, janko2019cross,saha2021empirical,zhu2021data}, have demonstrated their effectiveness in analyzing data obtained from accelerometers, gyroscopes, magnetometers, linear accelerometers, gravity, orientation, and ambient pressure sensors. Support Vector Machines (SVMs) are frequently utilized in this particular domain \cite{wu2018decision} and are frequently regarded as a standard against which deep learning methods are compared. 
Despite the performance superiority, these traditional machine learning methods often require the extraction of features in both the time and frequency domains, and they normally require a large amount of sensory data to extract those features.

Moreover, instead of using raw sensory data, traditional machine learning techniques are often applied for feature transformation. 
The Fast Fourier Transformation is frequently used for gleaning frequency domain information from sensory data \cite{janko2018new,janko2019cross,kalabakov2020tackling}, especially the time domain features. 
For example, Janko et al.~\cite{janko2018new} highlighted the role of expert knowledge in selecting frequency domain features. 
Specifically, they chose to examine three primary magnitudes: Energy, Entropy, and Binned distribution, along with Skewness and Kurtosis. 
Widhalm et al.~\cite{widhalm2018top} also incorporated autocorrelation as a key feature in their research.
The process of feature transformation requires not only expert knowledge but also additional efforts from domain experts.

In addition, the sequential signal measurements are summarized in statistics before feeding into a detection model \cite{janko2018new,janko2019cross,kalabakov2020tackling,zhu2021data,ren2021multiple}.
These statistical features include minimum, maximum, mean, variance, standard deviation, mean absolute deviation, autocorrelation, count of samples above or below the mean, and the average variance between consecutive data points. 
Whereas, these high-level statistics inevitably lose the detailed signal changes in consecutive measurements with a high frequency, \eg~\SI{100}{Hz}, which can be very helpful to capture motion patterns in different transportation modes.


\subsection{Deep learning methods}

In recent years, deep learning methods have garnered significant attention with notable improvements in accuracy for transportation mode detection.
The commonly used network architectures are, \eg~Convolutional Neural Networks (CNNs), Long Short-Term Memory (LSTM) model, and Gated Recurrent Unit(GRU). 
For example, \cite{fang2017learning,liang2019deep} employed CNNs with time-series inputs for transportation mode detection.
Thanks to the gating mechanisms designed for propagating information along the time axis, \cite{mishra2020sensors} combined LSTM with a CNN, achieving impressive mode detection performance on two distinct datasets\cite{mishra2020sensors}.
Similarly, GRU is also proven to be very efficient in mode detection problem \cite{zhu2020densenetx}.
Similar to LSTM and GRU, 1D convolutional layers have been demonstrated in handling time-series of mode detection, as demonstrated by 1D DenseNet proposed in \cite{zhu2019applying}.



Compared to traditional machine learning methods, even though some deep learning techniques do engage in feature extraction (e.g.\cite{gjoreski2020classical,zhu2020densenetx}), it is not always mandatory. For example, Choi et al.~\cite{choi2019embracenet}, although purportedly executing feature extraction, employs convolutional layers on raw data and feed them into an EmbraceNet.  
More commonly, a data-loader module is called to directly prepare the input from raw data.
This module includes functions to compute the magnitude of variables like acceleration, gyroscope, and magnetic field \cite{zhu2019applying}, jerk of the sensors \cite{tang2022}, and adapting the mobile phone's coordinate system to a global reference frame, among others \cite{gjoreski2020classical}.
In this way, signal expert knowledge is not necessarily required to pre-process on the raw data, which makes it considerably distinct from the feature extraction concept discussed earlier.
However, these above mentioned models have not fully explored a strategy to reduce the amount of sensory data and build a more efficient and effective model for transportation mode detection.
Therefore, in this paper, we design our own data-loader to prepare the input from raw sensory data and train an more lightweight end-to-end deep learning model for transportation mode detection.



\section{Mode Detection using Locomotion data}
\subsection{Problem formulation}
In this work, we leverage locomotion data collected from motion-based sensors for transportation mode detection.
Mathematically, the detection task is defined as given a frame of sensor measurements, denoted as $X_i$, in a certain reading rate, \eg~\SI{100}{Hz}, the detection 
function $f(X)$ is to recover the transportation mode $Y$.
Each frame of a certain length contains a set of sensor measurements, such as accelerometer, gyroscope, magnetometer, linear accelerometer, gravity, orientation, and ambient pressure.
Namely, $X_i = \{x_{i,1}^t, x_{i,2}^t, \dots, x_{i,j}^t, \dots\}_{t=1}^T$, where $j$ stands for one of the aforementioned sensor types and $t$ is the time step of a total length of $T$ time steps.
The objective function is to minimize the detection errors $\sum_{i=1}^{N}L(\hat{Y}_i, Y_i)$ between the prediction mode $\hat{Y}_i = f(X_i)$ and ground truth mode $Y_i$ across all the $N$ frame samples.

\subsection{Dataset and data-loader}
\label{section_data_preprocessing}
Concretely, we use the University of Sussex-Huawei Locomotion (SHL) dataset \cite{Gjoreski2018}.
It comprises eight primary modes, \ie~\textit{Still}, \textit{Walk}, \textit{Run}, \textit{Bike}, \textit{Car}, \textit{Bus}, \textit{Train}, and \textit{Subway}.
This dataset was collected by three users with their smartphones at four distinct positions, including the hand, hips, torso, and bag. 
It consists of the recorded measurements of various sensors, including the accelerometer, gyroscope, magnetometer, linear accelerometer, gravity, orientation (expressed as quaternions), and ambient pressure. 
Whereas, this study focuses on the utilization of accelerometer, magnetometer, and gyroscope  measurements for the purpose of transportation mode detection, with an emphasis on maintaining a low level of computing complexity. In total, this SHL dataset contains 753 hours of locomotion data, which has been appropriately labeled. The duration distributions of various modes are as follows: Car (88 hours), Bus (107 hours), Train (115 hours), Subway (89 hours), Walk (127 hours), Run (21 hours), Bike (79 hours), and Still (127 hours). 
The provided data is divided into distinct frames of one-minute duration, with the frames being randomly arranged to minimize temporal interdependence. 
Every frame comprises 6,000 samples, representing a duration of \SI{60}{s}, collected at a sampling rate of \SI{100}{Hz}. 

A subset of the SHL dataset known as \emph{SHL challenge 2018} was utilized for our detection task. Specifically, 271 hours were designated for training purposes, while 95 hours were allocated for testing purposes. At the sample frequency level, the training set has a total of 16,310 frames, resulting in a total training data size of $16310\times6000$ for each sensor. 
Similarly, the test set has a size of $5698\times6000$ for each sensor.
It should be noted that during the training process, we utilize one-hot encoding for the purpose of labeling. 
However, as a result of the data segmentation process, it is possible for a time window, \eg~\SI{60}{s}, to encompass multiple transportation modes. 
In order to resolve the controversy arising from the labels, the majority labeling policy is implemented, as described in reference \cite{gjoreski2020classical}.
Moreover, the impact of altering the window length on the model's performance was examined in accordance with the details provided in Section \ref{sub:frequency}.

Before introducing the proposed detection model, we first introduce the data-loader to facilitate an end-to-end training and detection.
Concretely, the data-loader provides a variety of functions to automatically process the data for the model, such as smoothing, downsampling, magnitude calculation, and jerk calculation.
\paragraph{Smoothing}
The received results from an IMU are acknowledged to be susceptible to bias and noise \cite{Titterton2005}. Smoothing is a commonly employed method utilized to mitigate the presence of random and undesired fluctuations within a dataset.
The algorithm employed in this study is a central moving average technique, specifically a Savitzky-Golay filter \cite{savitzky1964smoothing}. This filter calculates the average of an odd number of neighboring data points surrounding a given data item. The value of $m$, which represents the number of nearest neighbors, is predetermined and discussed in reference \cite{liang2019deep}. The averaged smoothing value is calculated in \eqref{eq:smoothing}.
\begin{equation}\label{eq:smoothing}
\bar{d}_{t} =\left\{
\begin{array}{ll}
\frac{{\sum }_{i=1}^{2t-1}{d}_{i}}{2{t}_{-1}} & t\in\left[1, \lfloor\frac{m}{2}\rfloor\right], \\
\frac{{\sum }_{i=t-(m/2)}^{t+(m/2)} d_{i}}{m} & t\in \left(\lfloor\frac{m}{2}\rfloor, T - \lfloor\frac{m}{2}\rfloor\right), \\
\frac{{\sum}_{i = 2t - T}^{T}d_{i}}{2(T-t)+1} & t \in \left[ T- \lfloor \frac{m}{2}\rfloor, T\right],
\end{array}
\right.
\end{equation}
where $t$ represents the $t$-th sample of a frame,  and $T$ is the total number of samples in a frame, and $d_{t}$ represents the $x$, $y$, and $z$ axis of accelerometer, gyroscope or magnetometer measurements, and $\bar{d}_{t}$ is the smoothed data. 

\paragraph{Downsampling}
Downsampling the dataset reduces its size, making it more manageable and decreasing the computing cost of the learning process. Mishra et al. \cite{mishra2020sensors} applied downsampling to decrease the sample frequency by calculating the average of the data, and in this paper we apply the similar strategy. The $p$-th data point after downsampling, $\hat{d}_{p}$, is obtained with \eqref{eq:downsaplinWg}. 
\begin{equation}\label{eq:downsaplinWg}
    \hat{d}_{p} = \frac{ {\sum}_{t}^{t+S-1} \bar{d}_i}{S}, 
\end{equation}
where $t= nS+1,  n = \{0, 1, 2, ..., \lfloor F/S \rfloor\}$, and $S$ is the downsampling size and $F$ represents the original sampling frequency. For example, in the SHL dataset, if sampling frequency $F$ is $100\,\si{Hz}$ and $S=2$, every two data points are averaged so the frequency after downsampling becomes $50\,\si{Hz}$.
It should be noted that in order to determine the optimal settings, we employ various downsampling frequencies and conduct a comparative analysis. Further information regarding this experiment can be found in Section~\ref{section_experiment}.

\paragraph{Magnitude calculation}
The direction of axes (\ie~$x$, $y$, and $z$) of a sensor measurements are determined by the phone coordinate system. 
However, the values alter drastically if the smartphone's orientation is not fixed. Consequently, applying magnitude seeks to eliminate the influence of orientation changes on the detection of the transport mode, as stated in \cite{liang2019deep, fang2017learning, iskanderov2020breaking}. 
The magnitude of each sensing data at data point $p$, $M_{p}$, is calculated as shown in \eqref{eq:magnitude}.
\begin{equation}\label{eq:magnitude}
M_{p} = \sqrt{ \hat{d}_{x,p}^{2} + \hat{d}_{y,p}^{2} + \hat{d}_{z,p}^{2}},
\end{equation}
where $\hat{d}_{x, p}$, $\hat{d}_{y, p}$ and $\hat{d}_{z, p}$ represent the data point from the $x$, $y$, and $z$ axes of one sensor from accelerometer, gyroscope or magnetometer after downsampling, respectively.

\paragraph{Jerk calculation}
Jerk originally represents the rate at which acceleration changes [18]. During abrupt movements, acceleration is not uniform, and its estimation can be aided by analyzing the jerk. This concept has frequently been used in GPS-based mode detection \cite{iskanderov2020breaking, dabiri2018inferring}, primarily due to its implications in safety scenarios like crucial driving actions and maintaining passenger stability in public transit \cite{dabiri2018inferring}. Our objective is to discern the advantages and understand the potential impact of estimating the jerk on transportation mode detection, as alluded to in \cite{antar2018comparative}. The same principle is used for gyroscope and magnetometer readings, helping calculate rotation rates between two distinct data points around a given axis. 
Specifically, $\vec{J}_{p}$ signifies the jerk of specific data points $\vec{\hat{d}}_{p+1}$ and $\vec{\hat{d}}_{p}$ for sensors falling in accelerometer, gyroscope or magnetometer. 
The equation to obtain jerk is shown in \eqref{eq:jerk}.
\begin{equation}\label{eq:jerk}
 \vec{J}_{p} = \frac{\vec{\hat{d}}_{p+1} - \vec{\hat{d}}_{p}}{\Delta t}
\end{equation}
where $p$ denotes the data point index and $\Delta t$ is the time difference between successive data point $p+1$ and $p$.

\subsection{Feature Pyramid biLSTM model}\label{section_model}
\begin{figure}[hbpt]
	\centering
	\includegraphics[width=\linewidth]{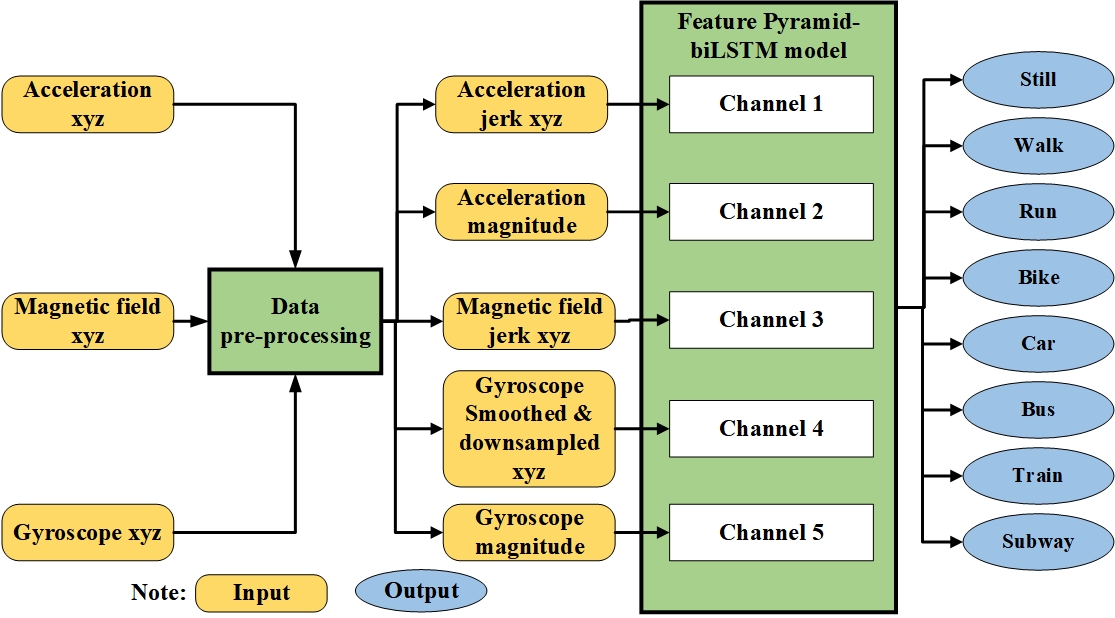}
	\caption{System architecture of Feature Pyramid biLSTM.}
	\label{fig:system_architecture}
\end{figure}

As illustrated in Figure~\ref{fig:system_architecture}, our proposed model is developed based on the CNN biLSTM model \cite{tang2022} and FPN \cite{lin2017featurepyramid}. 
The CNN biLSTM model evolves from Liang et al. \cite{liang2019deep}, which drew inspiration from AlexNet \cite{krizhevsky2012imagenet} and utilizes only linear acceleration as an input for transport mode detection. 
On the contrary, we incorporate the outputs of various convolutional layers into the biLSTM and fully-connected layers to facilitate prediction. 
This approach enables the utilization of abundant information from the shallow layer as well as the extracted features from the deeper layer, thereby further enhancing the model's capability. 
We refer to our model as the Feature Pyramid biLSTM model (FPbiLSTM).

FPbiLSTM takes as input the measurements from accelerometer, magnetometer, and gyroscope using the data-loader defined in Section \ref{section_data_preprocessing}.
Namely, they are jerk measurements from the $x$-, $y$-, and $z$-axis of acceleration, overall acceleration magnitude, jerk measurements from the $x$-, $y$-, and $z$-axis of the magnetic field, smoothed and down-sampled readings from the $x$-, $y$-, and $z$-axis of the gyroscope, and the gyroscope's overall magnitude. 
The original \emph{SHL challenge 2018} dataset consists of frames in a duration of \SI{60}{s} with a sampling frequency of $100\,\si{Hz}$. In the proposed model, a downsampling frequency of $20\,\si{Hz}$ is used.  
In the following, we explain each step of the model in detail.

\begin{figure}[hbpt]
	\centering
	\includegraphics[clip=true, trim=2pt 2pt 2pt 2pt, width=\linewidth]{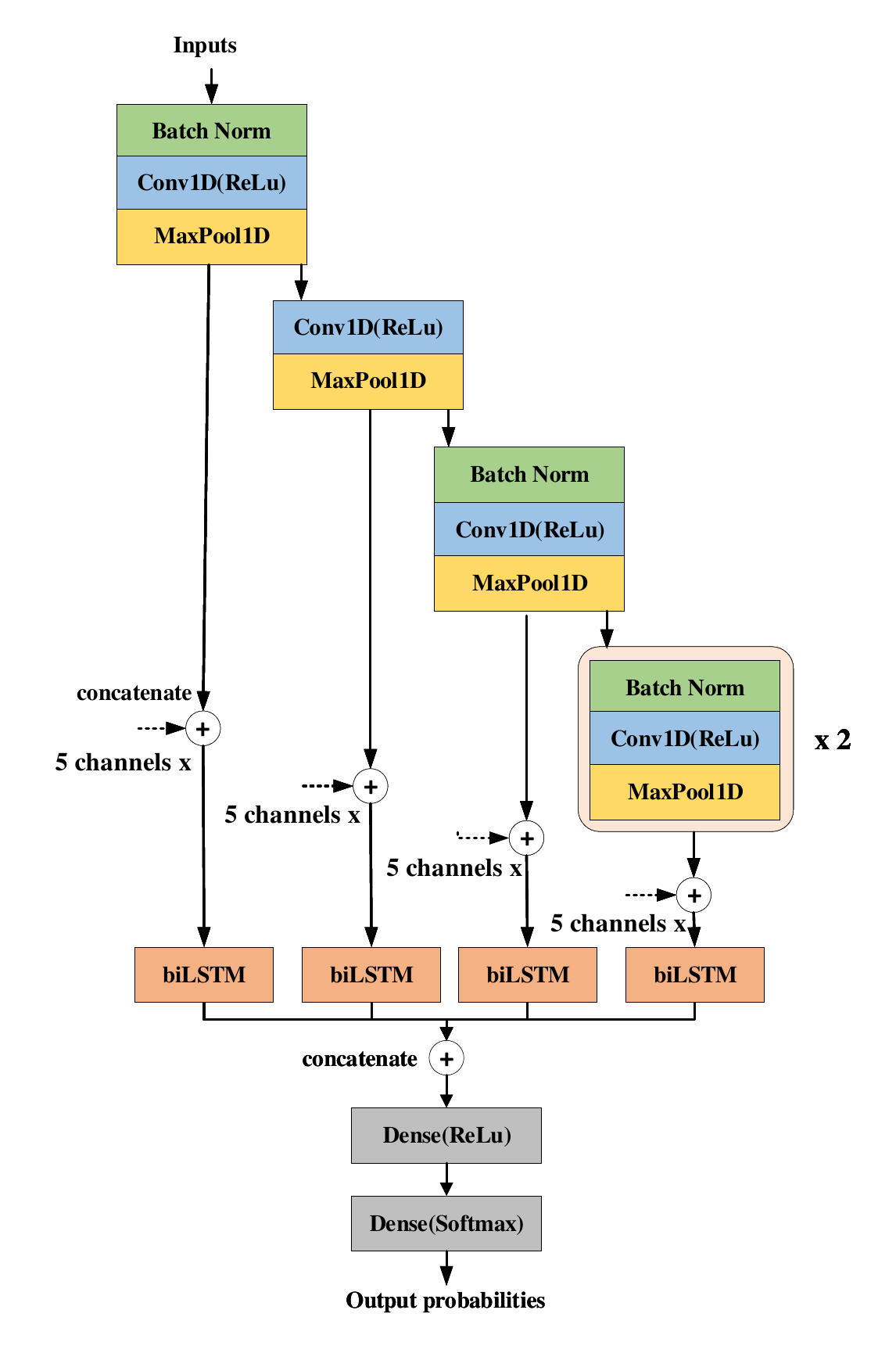}
	\caption{Architecture of Feature Pyramid biLSTM.}
	\label{fig:model_architecture}
\end{figure}

As shown in Figure \ref{fig:model_architecture}, first, the FPbiLSTM model processes each input through five separate channels to represent each input into a higher dimension separately.
Each channel's architecture and hyperparameters share the same features. 
For every channel, the filter count in the convolutional layers, commonly known as a filter bank, gradually increases, while the kernel size diminishes in the layers that follow. 
Each layer's filter banks have a stride size of 1. 
Due to the high frequency of sensor measurements, we use a relatively large kernel size to process the input.
Specifically, the inaugural convolutional layer consists of 32 filters with a kernel size of 15 each. The subsequent two layers feature 64 filters and a reduced kernel size of 10. The fourth and fifth convolutional layers incorporate 128 filters, each with a kernel size of 5. After every convolutional layer, a max-pooling layer is added, boasting a pooling size of 4 and a stride size of 2. 
To tackle the internal covariate shift and maintain stability during the model's training, batch normalization \cite{ito2018application} is applied before the first, third, fourth, and fifth convolutional layers. 
However, empirically, we found that introducing batch normalization before the second layer does not notably improve the network's efficacy.
Hence, it is not applied to the second layer.
At the end of the last convolutional layer, the channel-wise feature maps are concatenated along the channel axis to fuse them.

Consequently, we employ a biLSTM to learn the temporal information from the feature maps outputted by the above convolution layers.
In order to facilitate the learning of temporal information at different levels, we use skip connections to feed the feature maps to the biLSTM.
Concretely, the outputs of the first, second, third, and fifth max-pooling layers are fed into four biLSTM layers with a unit size of 128. 
This biLSTM model is composed of two LSTMs. 
They process the input in both the forward and backward directions, so as to capture additional contextual information and to yield superior performance compared to a unidirectional LSTM. 
A more detailed ablation study on the skip connections is provided in Section \ref{subsec:ablation}.

In the end, the learned information is fed to two stacked dense layers for computing the classification scores. 
The first layer has 128 units and the following layer has 8 units, matching the overall count of transportation modes. The outputs after the second dense layer followed by the Softmax activation represents the probability scores across different transportation methods. Finally, the mode indexed with the greatest score is identified as the anticipated transportation mode.

The selection of a suitable activation function is pivotal in the design of a neural network's architecture. For the hidden layers, the Rectified Linear Unit (ReLU) activation function was chosen due to its effectiveness in mitigating the vanishing gradients issue\cite{Xu2015}. For multi-class classification tasks, the Softmax function is used in the output layer to produce a set of values denoting the probability of each category. 

\vspace{3pt}
\noindent\textbf{Loss function.} The proposed model utilizes the Mean Squared Error (MSE) as the loss function, measuring the mean of squared differences. To reduce the loss with every epoch, the Adaptive Moment Estimation (Adam) optimizer is applied\cite{kingma2014adam}. Unlike the conventional stochastic gradient descent with a static learning rate, the Adam optimizer can dynamically modify the learning rate, thereby optimizing its efficiency. Lastly, to counteract overfitting on training data, introducing $L2$ regularization to the first dense layer proved to be effective.

\section{Experiment}\label{section_experiment}
\subsection{Experiment settings and evaluation metrics}
To facilitate the training of the network, the initial training dataset is partitioned into two subsets: a sub-training set and a sub-validation set. This partitioning is achieved by employing the Stratified Shuffle Split technique\cite{Brewer1999}, with a ratio of 90:10, thus the label distribution of the sub-training set and the sub-validation set is same. The test set utilized in this study is identical to the original SHL challenge dataset. In order to mitigate the issue of overfitting, the early stopping method with patience 5 is employed throughout the training phase, wherein the loss and accuracy on the sub-validation set are continuously monitored. The corresponding values of hyperparameters of the model are presented in Table~\ref{tb:parameter}. 

\begin{table}[!ht]
	\caption{Hyperparameters for the Feature Pyramid biLSTM model}
	\label{tb:parameter}
	\begin{center}
		\begin{tabular}{l|c}
			\toprule
			Hyperparameters& Value \\ \midrule
			L2 regularization & $0.001$ \\
			Learning rate & $0.0001$ \\  
			Minimum learning rate & $0.00001$ \\  
			Factor of reduced learning rate & $0.2$ \\  
			First order moment weight in Adam & $0.9$\\  
			Second order moment weight in Adam & $0.999$ \\ 
			Batch size & $50$ \\  
			\bottomrule 
		\end{tabular} 
	\end{center}
\end{table}

The model is developed using the Keras framework and trained on eight GTX 1080 Ti GPUs less than one hour. 
While the inference was computed on a single GTX 1080 Ti GPU. 
We reported more detailed computational performance in Tables \ref{tb:conv_layers} and \ref{tb:feature_pyramid}. 
In training, the model starts with randomized weights, leading to potential variations in the following evaluation metrics across different training sessions. We run the training process ten times and report the optimal results.

In order to assess and compare the performance of our proposed network with existing approaches, we employ accuracy, and the macro-averaged  $F_1$-score \cite{Dalianis2018} as indicated in \eqref{eq:f1}.
\begin{equation}\label{eq:f1}
F_1 = \frac{1}{C}\sum_{k=1}^{C} \frac{2 \cdot recall_{k}\cdot precision_{k}}{recall_{k}+precision_{k}}
\end{equation}
where $C$ is the number of transportation modes, $recall_{k}$ is the recall of class $k$, and $precision_{k}$ is the precision of class $i$.

The proposed model is designed as a sequence-to-one model, wherein the predicted class remains consistent inside a frame regardless of the number of samples present. Hence, it is imperative to acknowledge that the accuracy and  $F_1$-score represent the performance metrics on a per-frame basis. In line with the \emph{SHL challenge 2018}, where the performance per sample was reported, we also provide an evaluation of \emph{accuracy} and \emph{$F_1$-score per sample} for the purpose of model comparison. The analysis of the impact of the majority labelling policy is conducted in Section~\ref{subsection_influence_labelling}. 

\subsection{Results}\label{subsection_results}

\begin{figure}
	\centering
	\includegraphics[width=0.9\linewidth]{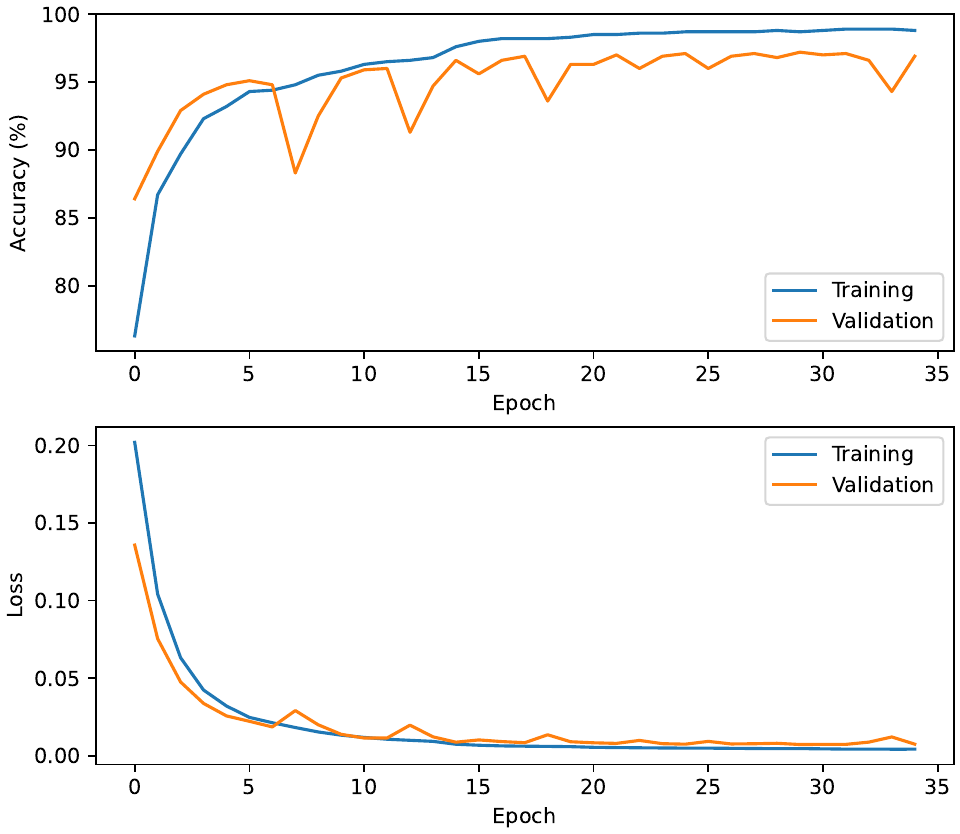}
	\caption{Learning curve.}
	\label{fig:learning_curve}
\end{figure}

By utilizing a time window of \SI{60}{s} and a downsampling frequency of \SI{20}{Hz}, 
the model achieves an accuracy of 95.1\% and a $F_{1}$-score of 94.7\%. 
Moreover, Figure~\ref{fig:learning_curve} depicts the learning curve. The initial epoch exhibits fluctuating validation accuracy and loss, which subsequently stabilize. The training process concludes at epoch 35 as a result of the use of early stopping. Based on the absence of an increasing trend in the validation loss curve, it is inferred that overfitting was mitigated throughout the training process.   

\unskip

\begin{table}[!ht]
\centering
	\caption{Confusion matrix with frequency $20\,\si{Hz}$ in time window $60\,s$}
	\setlength{\tabcolsep}{4pt}
	\label{tb:confusion_matrix}
	\begin{center}
		\begin{tabular}{|c|c|c|c|c|c|c|c|c|c|}
			\hline 
				\multicolumn{2}{|c|}{ }	& \multicolumn{8}{c|}{Predicted labels} \\
				\cline{3-10}
			\multicolumn{2}{|c|}{ }	&	Still	&	Walk	&	Run	&	Bike	&	Car	&	Bus	&	Train	&	Subway	\\
			\hline
			\multirow{8}{*}{\rotatebox{90}{Ground Truth}} 
            &	Still	&	923	&	6	&	0	&	1	&	5	&	15	&	9	&	2	\\
            &	Walk	&	9	&	719	&	2	&	0	&	0	&	0	&	1	&	0	\\
            &	Run	&	0	&	1	&	336	&	0	&	0	&	0	&	0	&	0	\\
            &	Bike	&	3	&	0	&	0	&	508	&	0	&	0	&	0	&	0	\\
            &	Car	&	10	&	0	&	0	&	0	&	1249	&	12	&	3	&	2	\\
            &	Bus	&	40	&	4	&	0	&	0	&	18	&	826	&	4	&	9	\\
            &	Train	&	41	&	2	&	0	&	0	&	13	&	4	&	541	&	46	\\
            &	Subway	&	7	&	0	&	0	&	0	&	0	&	0	&	19	&	308	\\
	   \hline																	
            &	Recall	&	96.0	&	98.4	&	99.7	&	99.4	&	97.9	&	91.7	&	83.6	&	92.2	\\
            &	Precision	&	89.4	&	98.2	&	99.4	&	99.8	&	97.2	&	96.4	&	93.8	&	83.9	\\
            &	$F_{1}$-score	&	92.6	&	98.3	&	99.6	&	99.6	&	97.5	&	94.0	&	88.4	&	87.9	\\
			\hline 
		\end{tabular} 
	\end{center}
\end{table}

Table~\ref{tb:confusion_matrix} presents the per-class confusion matrix, precision, recall, and $F_{1}$-score at a downsampling frequency of \SI{20}{Hz} and a time window of \SI{60}{s}. The $F_{1}$-scores for the categories of Walk, Run, Bike, and Car in the confusion matrix demonstrate a high level of effectiveness in classification (over 97\%). 
In contrast, the $F_{1}$-scores for the Train and Subway models exhibit considerably lower values. Similar observations of frequent mis-classification between Train and Subway modes can be found in previous studies analyzing the SHL dataset \cite{Gjoreski2018}.
Our conjecture is that the motion dynamics by train and subway are very similar, especially in the city environment when the trains and subways travel at similar speed. 
Extra information, such as location-based data, may be needed to distinguish between them.
We leave this aspect as our future work.

\begin{figure}[hbpt!]
	\centering
	\includegraphics[width=0.9\linewidth]{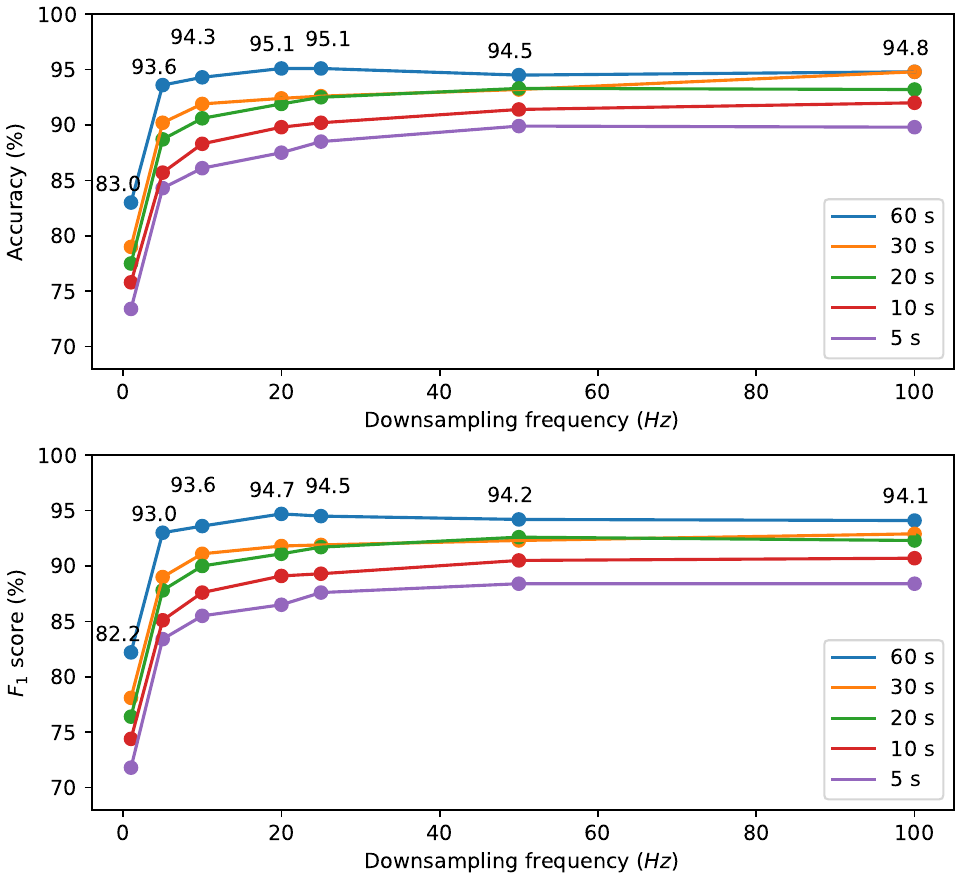}
	\caption{Evaluation on different time window and downsampling frequencies.}
	\label{fig:timewindow_downsampling}
\end{figure}

\subsection{Evaluation on time window and downsampling frequency}
\label{sub:frequency}
The proposed model was evaluated by altering the lengths of the time window and the downsampling rates. The time frame duration of \SI{60}{s} was initially studied, as specified in the SHL dataset. Following this, we conducted an evaluation of shorter durations for the window lengths, specifically \SI{30}{s}, \SI{20}{s}, \SI{10}{s}, and \SI{5}{s}. This choice was made due to the fact that $60\,s$ is divisible evenly by these durations. The frequencies used for downsampling evaluation included \SI{100}{Hz} (representing no downsampling), \SI{50}{Hz}, \SI{25}{Hz}, \SI{20}{Hz}, \SI{10}{Hz}, \SI{5}{Hz}, and \SI{1}{Hz}.

The prediction accuracy on the test set, with the specified window lengths and downsampling frequencies, is depicted in Figure~\ref{fig:timewindow_downsampling}. In our findings, there is a general trend of enhanced model performance as the downsampling frequency augments. However, an exception arises when employing a time window of \SI{60}{s}. Specifically, with this time window, the accuracy and $F_{1}$-scores for the \SI{50}{Hz} and \SI{100}{Hz} downsampling frequencies are inferior compared to those at \SI{20}{Hz} and \SI{25}{Hz} where the model obtains highest accuracy 95.1\% and $F_{1}$-score 94.7\%. One plausible explanation is that the extended time series may impede the model's feature extraction capability. Furthermore, exceedingly low downsampling frequencies might overly dilute the information within the data, leading to a degradation in model performance. It is noteworthy to observe that there is a substantial improvement in accuracy when downsampling frequencies transition from \SI{1}{Hz} to \SI{5}{Hz}, regardless of the length of the window. In scenarios with shorter time windows, the reduction in data duration and inherent information is associated with diminished performance. However, even within a 5-second time window at \SI{50}{Hz} and \SI{100}{Hz}, the model's accuracy nearly approaches an impressive 90\%. This underscores the model's efficacy in feature extraction and recognition.

\subsection{Feature contribution and selection}\label{section_feature}
In Section~\ref{section_data_preprocessing}, we delineated our data preprocessing approach. Subsequent to these preprocessing steps, we distilled three distinct features from each sensor: the accelerometer, gyroscope, and magnetometer. These features include the smoothed and downsampled data from the $x$-, $y$-, and $z$-axis, the magnitude, and the jerk. This section delves into an examination of the significance of these derived features.

Table~\ref{tb:feature_contribution} presents the performance metrics for each feature when employed as a solitary input, implying the model operates with just a single channel. The reported outcomes pertain to scenarios wherein the time window is fixed at \SI{60}{s}, the downsampling frequency is established at \SI{20}{Hz}, and a consistent random seed is deployed during execution to mitigate any fluctuations stemming from system randomness. As evidenced in the table, when considering acceleration data, the magnitude, when used as a solo input, achieves an accuracy of 89.7\% and an $F_{1}$-score of 88.9\%. Both in terms of accuracy and $F_{1}$-score, the jerk feature outperforms the $x$-, $y$-, and $z$-axis data by an approximate margin of 10 percent. For gyroscope data, both the smoothed \& downsampled $x$-, $y$-, and $z$-axis data and magnitude are markedly superior to the jerk feature. However, with respect to magnetic field data, jerk exhibits a notably enhanced performance compared to other features. Thus, based on our analyses, we posit that these five aforementioned features make the most substantial contributions to the model's performance outcomes.

\begin{table}[!ht]
\centering
\caption{Feature contribution from single sensor}
\label{tb:feature_contribution}
\small
\begin{tabular}{l|ccc|cc}
\toprule
\multirow{2}{*}{sensor}        & \multicolumn{3}{c|}{Feature} & \multirow{2}{*}{Accuracy} & \multirow{2}{*}{$F_{1}$-score} \\
                               & $x\,y\,z$   & $M_s$   & $J_s$   &                           &                           \\ \midrule
\multirow{3}{*}{Accelerometer} & $\surd$ &         &         &  76.7                    & 74.0                     \\
                               &         & $\surd$ &         &  \textbf{89.7}           & \textbf{88.9}            \\
                               &         &         & $\surd$ &  85.3                    &  82.9                   \\ \midrule
\multirow{3}{*}{Gyroscope}     & $\surd$ &         &         &  \textbf{82.2}           & \textbf{80.8}            \\
                               &         & $\surd$ &         &  81.6                    &  79.0                    \\
                               &         &         & $\surd$ &  77.3                    & 76.0                     \\ \midrule
\multirow{3}{*}{Magnetometer}  & $\surd$ &         &         &  62.9                    & 61.4                     \\
                               &         & $\surd$ &         &  61.6                    & 61.4                     \\
                               &         &         & $\surd$ &  \textbf{72.4}           & \textbf{74.5}            \\ \bottomrule
\end{tabular}
\end{table}

In light of the notable performance exhibited by individual features, it becomes imperative to investigate if amalgamating them might further enhance the outcomes. Table~\ref{tb:feature_combination} elucidates the results stemming from the integration of various features both within a single sensor and across different sensors. Regarding acceleration data, when magnitude and jerk are incorporated concurrently as dual channels, the model yields an accuracy of 90.3\% and an $F_{1}$-score of 88.6\%, surpassing the performance of any singular feature. Similarly, for gyroscope data, a synthesis of the smoothed \& downsampled $xyz$ with magnitude culminates in an accuracy of 86.5\%, which is notably superior to the 82.2\% achieved by the smoothed \& downsampled $xyz$ in isolation and the 81.6\% procured by the magnitude independently. These findings corroborate the proposition that strategic combinations can potentially amplify the model's efficacy.

Subsequently, we orchestrated a pairing of salient features from the accelerometer, gyroscope, and magnetometer, and then integrated these pairs with the prominent features from the trio of sensors. We discerned that specific combinations propelled the model's performance to even loftier heights.  Finally, with the combination of three sensors, the accuracy of the model reached 95.0\%. Although this marginally trails the apex accuracy of 95.1\% delineated in Section~\ref{subsection_results}, given the inherent randomness of the system, such a discrepancy is deemed within acceptable bounds.

\begin{table}[!ht]
\centering
\caption{Feature contribution from multiple sensors}
\label{tb:feature_combination}
\begin{tabular}{l|ccc|c|c|c|cc}
\toprule
\multirow{2}{*}{No.} & \multicolumn{3}{c|}{Sensor} & \multicolumn{3}{c|}{Feature} & \multirow{2}{*}{Accuracy} & \multirow{2}{*}{$F_{1}$-score} \\ 
  & A       & G       & M      & $xyz$      & $M$     & $J$   &                      &                      \\ \midrule
1 & $\surd$ &         &        &            & A         & A       & 90.3   & 88.6   \\
2 &         & $\surd$ &        & G          & G         &         & 86.5   & 84.8   \\
3 &  $\surd$ & $\surd$ &        & G          & A, G      & A       & 91.9   & 90.3   \\
4 &  $\surd$ &         &$\surd$ &            & A         & A, M    & 93.9   & 93.2   \\
5 &         & $\surd$ &$\surd$ & G          & G         & M       & 88.3   & 88.4   \\
6 &  $\surd$ & $\surd$ &$\surd$ & G          & A, G      & A, M    & 95.0   & 94.4   \\ \bottomrule
\multicolumn{9}{l}{\small A: Accelerometer; G: Gyroscope, M: Magnetometer}\\
\end{tabular}
\end{table}

\subsection{Ablation study}
\label{subsec:ablation}

Two ablation studies are conducted to ascertain the individual contributions of several factors to the performance of the model. The ablation study is employed to remove convolutional layers in order to get insight into their contribution as the depth of the model increases. Additionally, it investigates the connections between biLSTM and convolutional layers to assess the impact of the feature pyramid.

Table~\ref{tb:conv_layers} displays the results of ablating convolutional layers. Four tests were conducted. At the outset, the model exclusively preserved the initial convolutional layer, then augmenting the number of convolutional layers in a progressive manner, ultimately culminating in the inclusion of five convolutional layers, as per our proposed model. 
In this ablation experiment, our focus is solely on the contribution of the convolutional layer. 
Therefore, the connection relationship that forms the feature pyramid remains unchanged throughout the experiment. The table demonstrates a positive correlation between the number of convolutional layers and the performance of the model. Notably, the third convolutional layer has the most significant impact. The inclusion of a third convolutional layer in the model resulted in a significant improvement in accuracy, increasing from 78.6\% to 94.0\%. Subsequently, the task of enhancing accuracy encountered heightened challenges. Also, upon the inclusion of the fifth convolutional layer, a notable enhancement in accuracy was observed, with an increase of 0.9\%. This advancement can be considered substantial in the context of the study. Likewise, the quantity of parameters, duration of training, and floating point operations per second (FLOPS) all exhibit an upward trend as the model's depth increases, with the most pronounced variations observed specifically at the third convolutional layer.
It can be seen that with the addition of more convolutional layers, the complexity of the model increases.
While the model remains with a fast inference speed, \ie~the complete model (PFbiLSTM) takes only \SI{2.3}{ms} to detect the transportation mode for each frame.

\begin{table*}[!ht]
\centering
\caption{Ablation study on application of convolutional layers}
\label{tb:conv_layers}
\begin{tabular}{l|ccccc|c|cccc}
\toprule
\multirow{2}{*}{No.} & \multicolumn{5}{c|}{ Conv layers} & Accuracy & \# Params & Training time & Inference time & FLOPS \\
  & 1st  & 2nd  & 3th & 4th  & 5th & (\%) & (M) &  ($min$)  &  ($ms$)  &(G)                        \\ \midrule
1 & $\surd$ &  &  &  &  & 73.9 & 0.3 & 11.6 & 1.1 & 0.7 \\
2 & $\surd$ & $\surd$  &  & & & 78.6 &  0.9 & 10.0 & 1.6 & 6.8 \\
3 & $\surd$ & $\surd$ & $\surd$ &  &  & 94.0 & 1.6 & 21.1 & 1.9 & 13.0 \\ 
4 & $\surd$ & $\surd$ & $\surd$ & $\surd$ &  & 94.1 &  2.7  &    21.8   &              2.1                 & 16.1                     \\ \midrule
FPbiLSTM (ours) & $\surd$ & $\surd$ & $\surd$  & $\surd$  &  $\surd$  &  95.1  &  3.1  &   24.2    & 2.3  & 19.2          \\ \bottomrule

\end{tabular}
\end{table*}

The connection between the biLSTM layer and the convolutional layer is depicted in Table~\ref{tb:feature_pyramid} in order to explore the effect of feature pyramid. Except for the fourth convolutional layer and biLSTM, all the other convolutional layers are connected in our proposed model. To investigate the function of these connections, the ablation experiment interrupted some of the connections or added a fourth convolutional layer to the biLSTM. Cases No. 1-3 sequentially disconnect the initial, second, and third convolutional layers. When the connection between the three convolutional layers is disconnected and only the connection between the fifth convolutional layer is retained, the accuracy is as high as 94.9\%, which is significantly higher than the accuracy of Tang et al.\cite{tang2022} (91.2\%), indicating the improvement brought by feature selection.
After connecting all convolutional layers to the biLSTM layer, the accuracy falls to 94.5\%. As we propose, the optimal combination consists of concatenating the first, second, third, and fifth convolutional layers (an accuracy of 95.1\%). However, the number of parameters, training time and inference time increase considerably as the number of connections grows. Provides a hint for selecting model effectiveness and efficiency.

\begin{table*}[!ht]
\centering
\caption{Ablation study on feature pyramid}
\label{tb:feature_pyramid}
\begin{tabular}{l|ccccc|c|cccc}
\toprule
\multirow{2}{*}{No.} & \multicolumn{5}{c|}{Connection biLSTM \& Conv} & Accuracy & \# Params & Training time & Inference time & FLOPS \\
  & 1st  & 2nd  & 3th & 4th  & 5th & (\%) & (M) &  ($min$)  &  ($ms$)  &(G)                        \\ \midrule
1 &  & $\surd$ & $\surd$ &  & $\surd$ & 94.6 & 2.7 & 11.0 & 1.58 & 19.2 \\
2 & &  & $\surd$ & & $\surd$& 93.9 &  2.2 & 8.8 & 1.05 & 19.2 \\
3 &  &  &  &  & $\surd$ & 94.8 & 1.8 & 8.9 & 0.87 & 19.2 \\ 
4 & $\surd$ & $\surd$ & $\surd$ & $\surd$ & $\surd$ & 94.5 &  3.9  &    26.3   &    2.46                           & 19.2                     \\ \midrule
CNN-biLSTM\cite{tang2022}  &  &  &   &   &  $\surd$  &  91.2  &  2.1  &    10.1  &  1.6 &   57.7        \\ 
FPbiLSTM (ours)  & $\surd$ & $\surd$ & $\surd$  &   &  $\surd$  &  95.1  &  3.1  &   24.2    & 2.3  & 19.2          \\ \bottomrule

\end{tabular}
\end{table*}

\subsection{Influence of labeling policy}\label{subsection_influence_labelling}
The majority labeling strategy has been implemented in various studies, including \cite{gjoreski2020classical} and \cite{richoz2020transportation}, to handling transportation mode transitions in frames. Transition refers to the presence of multiple modes inside a frame resulting from the change of transportation modes. The labeling policy that assigns the majority label to the entire frame, considering it as a single transportation mode, is likely to have a detrimental impact on the performance of the model.

Table~\ref{tb:difference} presents the difference in $F_{1}$-scores between the per frame and per sample approaches. As the time window lowers, there is a drop in the fraction of transition frames among all frames in both the training set and test set. The time window is reduced and the transportation mode is more likely to appear in only one frame. As the ratio drops, there is a corresponding decrease in the difference between the $F_{1}$-score calculated per frame and per sample. When the time window duration is set to \SI{60}{s}, within the downsampling frequency range of \SI{10}{Hz} to \SI{100}{Hz}, the $F_{1}$-score per frame exhibits an average increase of over 0.5\% compared to the $F_{1}$-score per sample. When the duration of the time interval is set to \SI{30}{s}, the magnitude of this discrepancy is noticeably diminished. In the majority of cases, where the duration is shorter than \SI{30}{s}, the disparity is typically around 0.1\% and can be considered practically inconsequential.


\begin{table}[!ht]
	\caption{$F_{1}$-score difference between per frame and per sample}
	\setlength{\tabcolsep}{1.5pt}
	\label{tb:difference}
	\begin{center}
		\begin{tabular}{c|c|c|c|c|c|c|c|c|c}
			\toprule 			
Window & Training &	Test set  &	\multicolumn{7}{c}{$F_{1}$-score difference}												\\
size ($s$)	&	set ratio	&	ratio	&	\SI{1}{Hz}	&	\SI{5}{Hz}	&	\SI{10}{Hz}	&	\SI{20}{Hz}	&	\SI{25}{Hz}	&	\SI{50}{Hz}	&	\SI{100}{Hz}	\\
\midrule
60	&	3.99	&	3.97	&	0.29	&	0.53	&	0.53	&	0.51	&	0.52	&	0.63	&	0.50	\\
30	&	1.99	&	1.98	&	0.11	&	0.07	&	0.12	&	0.13	&	0.11	&	0.07	&	0.13	\\
20	&	1.33	&	1.33	&	0.05	&	0.09	&	0.12	&	0.14	&	0.16	&	0.14	&	0.13	\\
10	&	0.66	&	0.66	&	0.03	&	0.05	&	0.07	&	0.06	&	0.07	&	0.05	&	0.05	\\
5	&	0.33	&	0.33	&	0.01	&	0.02	&	0.02	&	0.02	&	0.02	&	0.00	&	0.02	\\

			\bottomrule 
		\end{tabular} 
	\end{center}
\end{table}

\subsection{Comparison with previous work}
In this section, the \emph{SHL challenge 2018} dataset was utilized by all previous studies participating in the comparison. In other words, the comparison did not encompass the utilization of additional data sources such as video and audio\cite{richoz2020transportation}. In order to maintain consistency in our analysis, we employed the $F_{1}$-score per sample as a benchmark when comparing our findings to those of prior studies.

Table~\ref{tb:compare} displays the comparison results with previous studies utilizing both traditional machine learning and deep learning techniques. Gjoreski et al. (ML+DL) \cite{gjoreski2020classical} demonstrates enhanced performance by the utilization of preprocessing techniques, incorporation of seven distinct sensor types, implementation of complicated feature extraction methods, and deep multimodal spectro-temporal fusion. namely, the authors of the study proposed a novel approach to classification, wherein they integrated a meta-model that combines deep learning and machine learning-based models. This was followed by a post-processing step involving the application of Hidden Markov Model (HMM) smoothing. By applying so, \cite{gjoreski2020classical} achieved the highest $F_{1}$-score 94.9\%, but with the model size \SI{500}{MB}. Nevertheless, our model exhibits a reduced weight and requires low resources for preprocessing, enabling it to categorize input data without necessitating subsequent post-processing. Upon utilizing data from the accelerometer, gyroscope, and magnetometer, the model exhibits a compact size of merely \SI{36}{MB}, while achieving a commendable $F_{1}$-score of 94.2\%. The size of the proposed model is greater than that of the Tang et al.'s model \cite{tang2022}, measuring 24MB. However, it has exhibited a notable enhancement in the $F_{1}$-score, demonstrating a 4\% increase. Consequently, the model's performance has experienced a significant advancement.

Table~\ref{tb:compare} also contrasts our technique with conventional machine learning approaches. The methodology employed by Janko et al. \cite{janko2018new} demonstrates a superior $F_{1}$-score (92.4\%), albeit with increased complexity resulting from the utilization of seven distinct sensor types, resulting in a model size of \SI{43}{MB}. Taking into account various elements such as the volume of sensing data, the utilization of a subset of the SHL dataset \cite{Gjoreski2018}, and the avoidance of intensive pre-processing for feature extraction, our model may be deemed lightweight and deserving of acknowledgment for effectively capturing essential characteristics of diverse transportation modes. 

\begin{table*}[!ht]
\centering
\caption{Comparison with state-of-the-art models}
\label{tb:compare}
\begin{tabular}{l|ccccccc|cc|c|c}
\toprule
\multirow{2}{*}{Method} & \multicolumn{7}{c|}{Sensing data} & \multirow{2}{*}{Input} & \multirow{2}{*}{Post processing} & Model size & \multirow{2}{*}{$F_{1}$-score} \\
                        & A  & G  & M & L  & g & O & Ap &       &  & (MB)                                &                           \\ \midrule
Random forest \cite{antar2018comparative} & $\surd$ & $\surd$ & $\surd$ & $\surd$ & $\surd$ & $\surd$ & $\surd$ & features & MV & 1122 & 87.5 \\
Multilayer perception \cite{widhalm2018top} & $\surd$ & $\surd$ & $\surd$ &&&& $\surd$ & features & HMM & 0.04 & 87.5 \\
XGBoost \cite{janko2018new} & $\surd$ & $\surd$ & $\surd$ & $\surd$ & $\surd$ & $\surd$ & $\surd$ & features & - & 43 &  92.4 \\ \midrule
DNN \cite{akbari2018hierarchical}                    & $\surd$ & $\surd$ & $\surd$ & $\surd$ & $\surd$ & $\surd$ & $\surd$   & features      & MV                               & 84 & 86.3                      \\
CNN \cite{ito2018application}                    & $\surd$ & $\surd$ &   &   &    &   &    & spectrogram      & -                                 &3 & 88.8                      \\
CNN+BiLSTM \cite{tang2022}             & $\surd$   &    & $\surd$  &   &   &   &     & raw data       & -                                & 24 & 90.4                      \\
ML+DL \cite{gjoreski2020classical}                  & $\surd$ & $\surd$ & $\surd$ & $\surd$ & $\surd$ & $\surd$ & $\surd$    & spectrogram + features      & HMM                              & 500 & 
 94.9                      \\
FPbiLSTM (ours)         & $\surd$ & $\surd$ & $\surd$ &   &   &   &   &  raw data      & -                                 & 36 &  94.2                     
 \\ \bottomrule
 \multicolumn{11}{l}{\small A: Accelerometer; G: Gyroscope, M: Magnetometer; L: Linear; g: Gravity; O: Orientation; Ap: Ambient pressure}\\
\multicolumn{11}{l}{\small MV: Majority Voting; HMM: Hidden Markov model}\\
\end{tabular}
\end{table*}

\section{Conclusion}

The ubiquity of smartphones equipped with IMUs offers a unique opportunity to harness a vast range of sensor data for numerous applications, most prominently, transportation mode detection. We aim to propose the Feature Pyramid biLSTM model, drawing its roots from both the FPN and the CNN biLSTM model, innovatively incorporating sensor data from accelerometers, gyroscopes, and magnetometers. This integration, despite its structural complexity, presented a compelling case of efficiency and reduced computational demand.

In our experiment using the \emph{SHL challenge 2018}, we were able to unravel the nuanced relationships between downsampling frequencies, time window lengths, and their impact on model performance. Evidently, while certain time windows and frequencies augmented the model's capabilities, exceedingly low downsampling frequencies and prolonged time series occasionally impeded effective feature extraction. Through extensive feature analysis, the study underscored the significance of specific data features and their potential enhancement when synergistically combined, a sentiment further echoed in our juxtaposition with different sensor amalgamations.

Benchmarking against prior research offered a testament to the model's prowess. While some techniques previously registered higher performance metrics, they often came laden with increased computational burdens, intensive preprocessing, or inflated model sizes. In contrast, our model epitomized efficiency and lightweight design, demanding minimal preprocessing while yielding commendable performance metrics. Notably, the success achieved through the utilization of only three sensors, with minor computational resources, underscores the model's potential for real-world applications, especially in environments constrained by resource availability or processing power.

\vspace{3pt}
\noindent\textbf{Limitation.} In this study, we found that the model has limited performance on distinguishing the modes between subway and train. 
This is because of their very similar motion patterns.
In our future work, we will explore additional information, such as the location-based sensory data from GNSS and GPS, to facilitate the mode detection. 
Moreover, we will explore more advanced deep learning models, such as Transformer~\cite{vaswani2017attention}, to increase the robustness and generalization on transportation mode detection by \eg~reducing the frame duration or testing on other datasets.





\bibliographystyle{IEEEtran}
 \bibliography{Reference}

\begin{thebibliography}{10}
\providecommand{\url}[1]{#1}
\csname url@samestyle\endcsname
\providecommand{\newblock}{\relax}
\providecommand{\bibinfo}[2]{#2}
\providecommand{\BIBentrySTDinterwordspacing}{\spaceskip=0pt\relax}
\providecommand{\BIBentryALTinterwordstretchfactor}{4}
\providecommand{\BIBentryALTinterwordspacing}{\spaceskip=\fontdimen2\font plus
\BIBentryALTinterwordstretchfactor\fontdimen3\font minus
  \fontdimen4\font\relax}
\providecommand{\BIBforeignlanguage}[2]{{%
\expandafter\ifx\csname l@#1\endcsname\relax
\typeout{** WARNING: IEEEtran.bst: No hyphenation pattern has been}%
\typeout{** loaded for the language `#1'. Using the pattern for}%
\typeout{** the default language instead.}%
\else
\language=\csname l@#1\endcsname
\fi
#2}}
\providecommand{\BIBdecl}{\relax}
\BIBdecl

\bibitem{Xiao2012}
Y.~Xiao, D.~Low, T.~Bandara, P.~Pathak, H.~B. Lim, D.~Goyal, J.~Santos,
  C.~Cottrill, F.~Pereira, C.~Zegras, and M.~Ben-Akiva, ``Transportation
  activity analysis using smartphones,'' in \emph{2012 IEEE Consumer
  Communications and Networking Conference (CCNC)}, 2012, pp. 60--61.

\bibitem{liang2019deep}
X.~Liang, Y.~Zhang, G.~Wang, and S.~Xu, ``A deep learning model for
  transportation mode detection based on smartphone sensing data,'' \emph{IEEE
  Transactions on Intelligent Transportation Systems}, vol.~21, no.~12, pp.
  5223--5235, 2019.

\bibitem{ito2018application}
C.~Ito, X.~Cao, M.~Shuzo, and E.~Maeda, ``Application of cnn for human activity
  recognition with fft spectrogram of acceleration and gyro sensors,'' in
  \emph{Proceedings of the 2018 ACM International Joint Conference and 2018
  International Symposium on Pervasive and Ubiquitous Computing and Wearable
  Computers}, 2018, pp. 1503--1510.

\bibitem{widhalm2018top}
P.~Widhalm, M.~Leodolter, and N.~Br{\"a}ndle, ``Top in the lab, flop in the
  field? evaluation of a sensor-based travel activity classifier with the shl
  dataset,'' in \emph{Proceedings of the 2018 ACM International Joint
  Conference and 2018 International Symposium on Pervasive and Ubiquitous
  Computing and Wearable Computers}, 2018, pp. 1479--1487.

\bibitem{tang2022}
Q.~Tang, K.~Jahan, and M.~Roth, ``Deep cnn-bilstm model for transportation mode
  detection using smartphone accelerometer and magnetometer,'' in \emph{2022
  IEEE Intelligent Vehicles Symposium (IV)}, 2022, pp. 772--778.

\bibitem{lin2017featurepyramid}
T.~Y. Lin, P.~Dollár, K.~Girshick, R .and~He, B.~Hariharan, and S.~Belongie,
  ``Feature pyramid networks for object detection,'' in \emph{Proceedings of
  the IEEE conference on computer vision and pattern recognition}, 2017, pp.
  2117--2125.

\bibitem{yu2014big}
M.-C. Yu, T.~Yu, S.-C. Wang, C.-J. Lin, and E.~Y. Chang, ``Big data small
  footprint: The design of a low-power classifier for detecting transportation
  modes,'' \emph{Proceedings of the VLDB Endowment}, vol.~7, no.~13, pp.
  1429--1440, 2014.

\bibitem{gjoreski2020classical}
M.~Gjoreski, V.~Janko, G.~Slapni{\v{c}}ar, M.~Mlakar,
  N.~Re{\v{s}}{\v{c}}i{\v{c}}, J.~Bizjak, V.~Drobni{\v{c}}, M.~Marinko,
  N.~Mlakar, M.~Lu{\v{s}}trek \emph{et~al.}, ``Classical and deep learning
  methods for recognizing human activities and modes of transportation with
  smartphone sensors,'' \emph{Information Fusion}, vol.~62, pp. 47--62, 2020.

\bibitem{janko2018new}
V.~Janko, N.~Re{\v{s}}{\c{c}}i{\c{c}}, M.~Mlakar, V.~Drobni{\v{c}}, M.~Gams,
  G.~Slapni{\v{c}}ar, M.~Gjoreski, J.~Bizjak, M.~Marinko, and M.~Lu{\v{s}}trek,
  ``A new frontier for activity recognition: the sussex-huawei locomotion
  challenge,'' in \emph{Proceedings of the 2018 ACM International Joint
  Conference and 2018 International Symposium on Pervasive and Ubiquitous
  Computing and Wearable Computers}, 2018, pp. 1511--1520.

\bibitem{janko2019cross}
V.~Janko, M.~Gjoreski, C.~M. De~Masi, N.~Re{\v{s}}{\v{c}}i{\v{c}},
  M.~Lu{\v{s}}trek, and M.~Gams, ``Cross-location transfer learning for the
  sussex-huawei locomotion recognition challenge,'' in \emph{Adjunct
  Proceedings of the 2019 ACM International Joint Conference on Pervasive and
  Ubiquitous Computing and Proceedings of the 2019 ACM International Symposium
  on Wearable Computers}, 2019, pp. 730--735.

\bibitem{choi2019embracenet}
J.-H. Choi and J.-S. Lee, ``Embracenet for activity: A deep multimodal fusion
  architecture for activity recognition,'' in \emph{Adjunct Proceedings of the
  2019 ACM International Joint Conference on Pervasive and Ubiquitous Computing
  and Proceedings of the 2019 ACM International Symposium on Wearable
  Computers}, 2019, pp. 693--698.

\bibitem{zhu2020densenetx}
Y.~Zhu, H.~Luo, R.~Chen, F.~Zhao, and L.~Su, ``Densenetx and gru for the
  sussex-huawei locomotion-transportation recognition challenge,'' in
  \emph{Adjunct Proceedings of the 2020 ACM International Joint Conference on
  Pervasive and Ubiquitous Computing and Proceedings of the 2020 ACM
  International Symposium on Wearable Computers}, 2020, pp. 373--377.

\bibitem{kalabakov2020tackling}
S.~Kalabakov, S.~Stankoski, N.~Re{\v{s}}{\v{c}}i{\v{c}}, I.~Kiprijanovska,
  A.~Andova, C.~Picard, V.~Janko, M.~Gjoreski, and M.~Lu{\v{s}}trek, ``Tackling
  the shl challenge 2020 with person-specific classifiers and semi-supervised
  learning,'' in \emph{Adjunct Proceedings of the 2020 ACM International Joint
  Conference on Pervasive and Ubiquitous Computing and Proceedings of the 2020
  ACM International Symposium on Wearable Computers}, 2020, pp. 323--328.

\bibitem{qin2019toward}
Y.~Qin, H.~Luo, F.~Zhao, C.~Wang, J.~Wang, and Y.~Zhang, ``Toward
  transportation mode recognition using deep convolutional and long short-term
  memory recurrent neural networks,'' \emph{IEEE Access}, vol.~7, pp.
  142\,353--142\,367, 2019.

\bibitem{fang2017learning}
S.-H. Fang, Y.-X. Fei, Z.~Xu, and Y.~Tsao, ``Learning transportation modes from
  smartphone sensors based on deep neural network,'' \emph{IEEE Sensors
  Journal}, vol.~17, no.~18, pp. 6111--6118, 2017.

\bibitem{saha2021empirical}
P.~Saha, M.~M. Alam, M.~I. Tapotee, S.~B. Baray, and M.~A.~R. Ahad, ``An
  empirical approach for human locomotion and transportation recognition from
  radio data,'' in \emph{Adjunct Proceedings of the 2021 ACM International
  Joint Conference on Pervasive and Ubiquitous Computing and Proceedings of the
  2021 ACM International Symposium on Wearable Computers}, 2021, pp. 390--395.

\bibitem{balabka2021human}
D.~Balabka and D.~Shkliarenko, ``Human activity recognition with automl using
  smartphone radio data,'' in \emph{Adjunct Proceedings of the 2021 ACM
  International Joint Conference on Pervasive and Ubiquitous Computing and
  Proceedings of the 2021 ACM International Symposium on Wearable Computers},
  2021, pp. 346--352.

\bibitem{munoz2023advanced}
E.~Munoz~Diaz, J.~M. Rubio~Hernan, F.~Jurado~Romero, A.~Karite,
  A.~Vervisch-Picois, and N.~Samama, ``Advanced smartphone-based identification
  of transport modes: Resilience under gnss-based attacks,'' \emph{Future
  Transportation}, vol.~3, no.~2, pp. 568--583, 2023.

\bibitem{antar2018comparative}
A.~D. Antar, M.~Ahmed, M.~S. Ishrak, and M.~A.~R. Ahad, ``A comparative
  approach to classification of locomotion and transportation modes using
  smartphone sensor data,'' in \emph{Proceedings of the 2018 ACM International
  Joint Conference and 2018 International Symposium on Pervasive and Ubiquitous
  Computing and Wearable Computers}, 2018, pp. 1497--1502.

\bibitem{zhu2021data}
Y.~Zhu, H.~Luo, S.~Guo, and F.~Zhao, ``Data mining for transportation mode
  recognition from radio-data,'' in \emph{Adjunct Proceedings of the 2021 ACM
  International Joint Conference on Pervasive and Ubiquitous Computing and
  Proceedings of the 2021 ACM International Symposium on Wearable Computers},
  2021, pp. 423--427.

\bibitem{wu2018decision}
J.~Wu, A.~Akbari, R.~Grimsley, and R.~Jafari, ``A decision level fusion and
  signal analysis technique for activity segmentation and recognition on smart
  phones,'' in \emph{Proceedings of the 2018 ACM International Joint Conference
  and 2018 International Symposium on Pervasive and Ubiquitous Computing and
  Wearable Computers}, 2018, pp. 1571--1578.

\bibitem{ren2021multiple}
Y.~Ren, ``Multiple tree model integration for transportation mode
  recognition,'' in \emph{Adjunct Proceedings of the 2021 ACM International
  Joint Conference on Pervasive and Ubiquitous Computing and Proceedings of the
  2021 ACM International Symposium on Wearable Computers}, 2021, pp. 385--389.

\bibitem{mishra2020sensors}
R.~Mishra, A.~Gupta, H.~P. Gupta, and T.~Dutta, ``A sensors based deep learning
  model for unseen locomotion mode identification using multiple semantic
  matrices,'' \emph{IEEE Transactions on Mobile Computing}, 2020.

\bibitem{zhu2019applying}
Y.~Zhu, F.~Zhao, and R.~Chen, ``Applying 1d sensor densenet to sussex-huawei
  locomotion-transportation recognition challenge,'' in \emph{Adjunct
  Proceedings of the 2019 ACM International Joint Conference on Pervasive and
  Ubiquitous Computing and Proceedings of the 2019 ACM International Symposium
  on Wearable Computers}, 2019, pp. 873--877.

\bibitem{Gjoreski2018}
H.~Gjoreski, M.~Ciliberto, L.~Wang, F.~J. Ordonez~Morales, S.~Mekki,
  S.~Valentin, and D.~Roggen, ``The university of sussex-huawei locomotion and
  transportation dataset for multimodal analytics with mobile devices,''
  \emph{IEEE Access}, vol.~6, pp. 42\,592--42\,604, 2018.

\bibitem{Titterton2005}
D.~Titterton and J.~Weston, \emph{Strapdown inertial navigation technology -
  2nd edition}.\hskip 1em plus 0.5em minus 0.4em\relax The institution of
  Engineering and Technology, London, UK and The American Institute of
  Aeronautics, Virginia, USA, 2004.

\bibitem{savitzky1964smoothing}
A.~Savitzky and M.~J. Golay, ``Smoothing and differentiation of data by
  simplified least squares procedures.'' \emph{Analytical chemistry}, vol.~36,
  no.~8, pp. 1627--1639, 1964.

\bibitem{iskanderov2020breaking}
J.~Iskanderov and M.~A. Guvensan, ``Breaking the limits of transportation mode
  detection: Applying deep learning approach with knowledge-based features,''
  \emph{IEEE Sensors Journal}, vol.~20, no.~21, pp. 12\,871--12\,884, 2020.

\bibitem{dabiri2018inferring}
S.~Dabiri and K.~Heaslip, ``Inferring transportation modes from gps
  trajectories using a convolutional neural network,'' \emph{Transportation
  research part C: emerging technologies}, vol.~86, pp. 360--371, 2018.

\bibitem{krizhevsky2012imagenet}
A.~Krizhevsky, I.~Sutskever, and G.~E. Hinton, ``Imagenet classification with
  deep convolutional neural networks,'' \emph{Advances in neural information
  processing systems}, vol.~25, 2012.

\bibitem{Xu2015}
B.~Xu, N.~Wang, T.~Chen, and M.~Li, ``Empirical evaluation of rectified
  activations in convolutional network,'' \emph{CoRR}, vol. abs/1505.00853,
  2015.

\bibitem{kingma2014adam}
D.~P. Kingma and J.~Ba, ``Adam: A method for stochastic optimization,''
  \emph{3rd International Conference on Learning Representations, ICLR}, 2015.

\bibitem{Brewer1999}
K.~R.~W. Brewer, ``Design‐based or prediction‐based inference? stratified
  random vs stratified balanced sampling,'' \emph{International Statistical
  Review}, vol.~67, pp. 35--47, 1999.

\bibitem{Dalianis2018}
H.~Dalianis, \emph{Clinical Text Mining: Secondary Use of Electronic Patient
  Records}.\hskip 1em plus 0.5em minus 0.4em\relax Springer International
  Publishing, 2018, ch. Evaluation Metrics and Evaluation, pp. 45--53.

\bibitem{richoz2020transportation}
S.~Richoz, L.~Wang, P.~Birch, and D.~Roggen, ``Transportation mode recognition
  fusing wearable motion, sound, and vision sensors,'' \emph{IEEE Sensors
  Journal}, vol.~20, no.~16, pp. 9314--9328, 2020.

\bibitem{akbari2018hierarchical}
A.~Akbari, J.~Wu, R.~Grimsley, and R.~Jafari, ``Hierarchical signal
  segmentation and classification for accurate activity recognition,'' in
  \emph{Proceedings of the 2018 ACM International Joint Conference and 2018
  International Symposium on Pervasive and Ubiquitous Computing and Wearable
  Computers}, 2018, pp. 1596--1605.

\bibitem{vaswani2017attention}
A.~Vaswani, N.~Shazeer, N.~Parmar, J.~Uszkoreit, L.~Jones, A.~N. Gomez,
  {\L}.~Kaiser, and I.~Polosukhin, ``Attention is all you need,''
  \emph{Advances in neural information processing systems (NeurIPS)}, vol.~30,
  2017.

\end{thebibliography}





\end{document}